\pdfoutput=1

\documentclass[11pt]{article}

\usepackage[final]{acl}

\usepackage{times}
\usepackage{latexsym}

\usepackage[T1]{fontenc}

\usepackage[utf8]{inputenc}

\usepackage{microtype}

\usepackage{inconsolata}

\usepackage{graphicx}
\usepackage[svgnames,dvipsnames,table]{xcolor}
\usepackage{makecell}
\usepackage{multirow}
\usepackage{tikz}
\usetikzlibrary{shapes.symbols,shapes,arrows,shapes.geometric}

\usepackage[table]{xcolor}
\usepackage{amsmath}
\usepackage{tabularx}
\newcolumntype{Y}{>{\centering\arraybackslash}X}
\usepackage{booktabs}

\usepackage{tablefootnote}

\newcommand{\ovalcell}[2][ForestGreen!30]{%
    \tikz[baseline=(text.base)]{%
        \node[fill=#1, rounded corners, inner sep=2pt] (text) {#2};%
    }%
}

\DeclareRobustCommand{\AdvArrow}{\begin{tikzpicture}[baseline=-0.3em] \node[shape=single arrow,draw,rotate=90,single arrow head extend=0.2em,inner ysep=0.2em,transform shape,line width=0.05em,top color=DarkOrchid,bottom color=DarkOrchid!50!black] (X){}; \end{tikzpicture}}

\DeclareRobustCommand{\DisArrow}{\begin{tikzpicture}[baseline=-0.3em] \node[shape=single arrow,draw,rotate=270,single arrow head extend=0.2em,inner ysep=0.2em,transform shape,line width=0.05em,top color=ForestGreen,bottom color=ForestGreen!50!black] (X){}; \end{tikzpicture}}

\title{Identifying Bias in Machine-generated Text Detection}

\author{Kevin Stowe, Svetlana Afanaseva, Rodolfo Raimundo, Yitao Sun, Kailash Patil \\ Pindrop \\ \texttt{\{kevin.stowe, safanaseva, rraimundo, ysun, kpatil\}@pindrop.com}}

\begin{document}

\interfootnotelinepenalty=10000

\maketitle
\begin{abstract}
The meteoric rise in text generation capability has been accompanied by parallel growth in interest in machine-generated text detection: the capability to identify whether a given text was generated using a model or written by a person. While detection models show strong performance, they have the capacity to cause significant negative impacts. We explore potential biases in English machine-generated text detection systems. We curate a dataset of student essays and assess 16 different detection systems for bias across four attributes: gender, race/ethnicity, English-language learner (ELL) status, and economic status. We evaluate these attributes using regression-based models to determine the significance and power of the effects, as well as performing subgroup analysis. We find that while biases are generally inconsistent across systems, there are several key issues: several models tend to classify disadvantaged groups as machine-generated, ELL essays are more likely to be classified as machine-generated, economically disadvantaged students' essays are less likely to be classified as machine-generated, and non-White ELL essays are disproportionately classified as machine-generated relative to their White counterparts. Finally, we perform human annotation and find that while humans perform generally poorly at the detection task, they show no significant biases on the studied attributes.
\end{abstract}

\section{Introduction}
    With the rise in the usage of generative AI systems, there is a growing need to distinguish content generated by a model from human-written content. To this end, there has been an explosion of research into machine-generated text detection\footnote{Also referred to as "deepfake" or "LLM-generated" text detection.}: identifying content that has been automatically generated by generative systems (e.g., large language models). These systems span detection of video, audio, and text-based generation, and are becoming essential tools for many practical scenarios where guidelines require human-written content.

    It is important to consider the practical implications of machine-generated text detection systems. In this work, we assess the potential for bias in these detection systems. There is substantial potential for harm in systems that erroneously flag content as automatically generated, especially if these systems exhibit bias toward disadvantaged populations. This could lead to rejection of genuine work, reduction of visibility, and unfair allocation of resources. Consider student essays, where inaccurate false positives could lead to harmful consequences for students, or content moderation tools, where human perspectives and representation could be unfairly filtered or silenced.
    
    To assess bias in machine-generated text detection systems, we systematically explore publicly available systems, analyzing their potential to unfairly classify human-written text as machine-generated. We curate a dataset of human-written texts and explore a series of publicly available models for potential biases across several dimensions: race, gender, age, ELL status, and economic status.  We pursue the following research questions:

    \begin{enumerate}
        \itemsep0em 
        \item \textbf{Do machine-generated text detection systems exhibit bias?} We are particularly interested in bias across dimensions of gender, race, age, ELL status, and economic status. We perform regression analysis, evaluating each attribute along with potential confounds.
        \item \textbf{Which subgroups are likely to be impacted?} We analyze each of the 16 possible subgroups, evaluating performance compared to the general population.
        \item \textbf{Do humans exhibit the same biases?} We have expert human annotators attempt to classify text as machine-generated or human-written, similarly evaluating their results for potential biases.
    \end{enumerate}
    
We find that while many English language learners texts are classified at a significantly higher rate than native speakers, these results are inconsistent, and models vary greatly in their respective biases. Several models tend to misclassify texts from disadvantaged groups as machine-generated, while other trained and zero-shot models vary. Across all attributes, we find that economic disadvantage serves as a negative indicator: essays from students with no economic disadvantage are classified as machine-generated more often across most systems. Subgroup analysis reveals more significant issues: non-White ELL essays are much more likely to be classified as machine-generated than White ELL essays, with higher incidence for men than for women. Finally, we find that human experts, while generally poor at the task (with accuracy ranging from 0.449 to 0.526), exhibit no significant biases based on the attributes studied.

\section{Background}
Our goal is to identify situations in which machine-generated text detection models make consistent mistakes with regard to certain attributes, violating group fairness \cite{czarnowska-2021}, which can lead to representational harms (e.g., protected groups misrepresented as abusing generative tools) and allocational harms (e.g., writing of protected groups being disqualified, censored, or minimized due to unfair model performance). We adopt their terminology: we analyze four \textbf{sensitive attributes}: gender, race/ethnicity, ELL status, and economic status. Within these attributes we define \textbf{advantaged/disadvantaged} groups (see Section \ref{sec:attributes}).\footnote{We use "disadvantaged" rather than "protected" as only two of our attributes, gender and race/ethnicity, are considered protected in the United States.} Our evaluation framework is based on \newcite{dayanik-2022}, who outline a method for identifying bias across multiple attributes in NLP problems.

\subsection{Machine-generated Text Detection}
Recent advances in generative AI have had many benefits, but understanding whether text has been written by a large language model (LLM) or a human is often essential. Domains such as news, where generative models can be used to spread misinformation \cite{hanley-2024,pan-2023}, education, where the use of generative AI in assisting students is under scrutiny \cite{meyer-2023}, and fraud, where generative models are being used to perpetrate scams \cite{romero-2025} highlight the need for accurate detection of machine-generated text.

The power of generative models has been accompanied by increased interest in detection systems (including a workshop focused specifically on this task \cite{genaidetect-2025}). There have been a wide variety of systems proposed, including feature-based methods, fine-tuned models, and zero-shot systems. For a comprehensive survey of recent datasets and methods, see \newcite{wu-2025}.

\subsection{Bias in Models}
Bias has been extensively studied in deepfake detection systems across domains of video and audio \cite{wang-2025,katamneni-2024,ju-2024}, and as a ubiquitous aspect of natural language processing \cite{bartl-2025,stanczak-2021,blodgett-2020}. Recent work has shown that humans exhibit significant biases when attempting to identify deepfake social media profiles \cite{mink-2024}. However, there remains relatively little work concerning the bias of machine-generated text detection systems.

\newcite{jung-2025} explores this topic in considerable depth, but focuses only on text length and stylistic personality as biases, which excludes disadvantaged groups. \newcite{liang-2023} explore bias in machine-generated text detection against non-native English speakers, claiming that seven major detectors perform significantly worse on non-native English speakers than native speakers. However, they do not indicate which models were used, and they report results on a limited dataset of only 179 student essays. \newcite{verma-2024} echo this result for English learner data, but note that it was unclear whether the differences were due to language or the length of the documents.

To address this gap in our understanding, we evaluate a suite of machine-generated text detection systems on a large corpus of student essays, evaluating performance for bias across four attributes: gender, race, ELL status, and economic status. We provide a thorough analysis of possible confounding factors as well as exploring subgroup differences. To our knowledge, this is the first work to perform a rigorous analysis of bias in machine-generated text detection.

\begin{table*}[t!]
\small
\begin{tabular}{lc|cc|cc|cc|cc}
    \toprule
    & & \multicolumn{2}{c|}{\textbf{Gender}} &  \multicolumn{2}{c|}{\textbf{Race/Ethnicity}} &  \multicolumn{2}{c|}{\textbf{ELL}} &  \multicolumn{2}{c}{\textbf{Economic Disadvantage}} \\
    & Total & Male & Female & White & Non-White & No & Yes & No & Yes \\
    \midrule
   \textsc{persuade v2.0} & 24695 & 12074 & 12621 & 11282 & 13413 & 22451 & 2244 & 11003 & 13692\\
    \textsc{asap v2.0} & 24728 & 12498 & 12230 & 9841 & 14887 & 20991 & 3737 & 7933 & 16795\\
    \textsc{ellipse} & 6482 & 3636 & 2846 & 471 & 6011 & 0 & 6482 & 1974 & 4508\\
    \midrule
    Combined (clean) & 41743 & 21277 & 20466 & 15078 & 26665 & 31079 & 10664 & 18188 & 23555 \\
    \bottomrule
    \end{tabular}
\caption{Counts for each attribute/group in our combined corpus.} 
\label{tab:dataset}
\end{table*}

\section{Evaluation Datasets}
    \label{sec:data}
    
    To evaluate potential bias in machine-generated detection systems, we use three datasets. These are \textsc{persuade-v2.0} \cite{crossley-2024a}, \textsc{asap-v2.0} \cite{crossley-2025}, and \textsc{ellipse} \cite{crossley-2024b}. These datasets all consist of persuasive essays written by 6th to 12th grade students in the United States, containing demographic information about race, gender, English-language-learner status, and economic status of the writers. The \textsc{asap-v2.0} and \textsc{ellipse} datasets extend the \textsc{persuade-v2.0} dataset: \textsc{asap-v2.0} adds 12k new samples and fills in demographic gaps, while the \textsc{ellipse} dataset adds new samples focusing on English language learners. We combine these datasets into a single, cleaned version, removing duplicate texts as well as instances where any demographic information is missing; details are in Table \ref{tab:dataset}. 

\subsection{Sensitive Attributes}
\label{sec:attributes}

    \paragraph{Gender:} The corpora contain gender as a binary attribute (male/female). We use these labels, with male considered the advantaged class and female as the disadvantaged class: machine learning models are known for biased performance on female data \cite{bartl-2025}. We recognize this binary labeling prohibits proper study and representation of other potential gender labels. Our setup mirrors previous work in which datasets contain only binary labels out of necessity \cite{biester-2025,plaza-2024,savoldi-2021}, and we continue with the understanding that this labeling system contains an inherent risk of misrepresenting bias, erasure, and other representational and allocational harms with regard to non-binary genders \cite{stanczak-2021,dev-2021}.

    \paragraph{Race/ethnicity:} The dataset contains six different labels for race/ethnicity. For our initial analysis, we compress these into two groups: White and non-White: for this attribute, this reflects the majority group (White) and the minority group (non-White) \cite{epi-2022}. This posits White as the advantaged class and non-White disadvantaged, but there are substantial differences between subgroups: we provide further exploration of the differences between these in Section \ref{sec:race}. We note such fixed categorization schemes can serve to entrench inequalities, and that racial divisions are a product of social contexts \cite{field-2021,hanna-2020}. Our goal is to examine potential biases in these models, but we stress our analysis inherits potential pitfalls from this labeling system.

    \paragraph{English-language learner (ELL) status:} The corpora make a binary distinction between ELL and non-ELL students; we consider ELL the disadvantaged attribute and non-ELL the advantaged.

    \paragraph{Economic status:} The corpora define two economic statuses: not disadvantaged (the "advantaged" group) and disadvantaged.

\section{Models}
    There are many options for machine-generated text detection models: we focus on an array of zero-shot and pretrained models.\footnote{Model implementation details provided in Appendix \ref{app:systems}.}

\subsection{Zero-shot Models}
    We split zero-shot models into GPT-based and non-GPT-based systems. All are suitable for zero-shot detection: they can score an independent text without training or other context. While not necessarily tuned to specifically detect GPT-based generation, the GPT-based models are clustered together as their dependence on OpenAI models makes them somewhat harder to inspect, and their behavior may change as access to these models changes. The models used are Ghostbuster \cite{verma-2024} and Glimpse \cite{bao-2025}. For non-GPT models, we utilize Fast-DetectGPT \cite{bao-2023} and Binoculars \cite{hans-2024}, which rely on extracting features from an underlying transformer-based language model, as well as Zippy \cite{thinkst-2023}, which uses compression-based methods.

\subsection{Trained Models}
    These are publicly available models that have been trained through varying methods on various datasets. We use BiScope \cite{guo-2024-1}, which has four variants (Yelp, Arxiv, Essay, and Creative) based on training data, and DeTeCtive \cite{guo-2024}, which has four variants of which we use three (MAGE, M4GT, TuringBench), as the fourth \textsc{outfox} variant overlaps with our evaluation data.

    Another subset consists of fine-tuned versions of transformer models. They have either been tuned for a specific task (Desklib \cite{desklib-2025} and e5-lora \cite{dugan-2024} are optimized for the RAID benchmark\footnote{\url{https://raid-bench.xyz/leaderboard}; at the time of writing, these are the two top-performing, publicly available systems.}), or are designed to be generally applicable for machine-generated text detection (RADAR \cite{hu-2023}).

\section{Benchmarking}
    We start by benchmarking the models to better understand their overall performance, and then explore potential biases on human-written corpora. To benchmark our models, we utilize the \textsc{outfox} dataset \cite{koike-2024}. This dataset combines human-written texts from the \textsc{persuade-v2.0} corpus with three machine-generated samples for each human-written text. This corpus comes from the same source, matching the domain, style, and tone of our human-written evaluation corpus.

    We evaluate each model, reporting precision, recall, F1 score, and area under the receiver operating characteristics (AUROC) in Table \ref{tab:benchmark}. These metrics provide a broad overview of performance: they have different implications for different use cases, with precision minimizing false positives, recall maximizing coverage, and AUROC providing a balanced view across thresholds.  We convert model scores into binary classification by identifying the threshold that optimizes equal error rate (EER) over a validation set of 1000 samples. We then use this threshold to make predictions, considering a sample machine-generated if the score provided by the model exceeds this threshold. This result is strictly improved F1 scores while keeping constant AUROC. 

    \begin{table}[t!]
\small
\setlength\tabcolsep{3.5pt}

\begin{tabular}{l|ccc|c}
\toprule
\textbf{Model} &  \textbf{Prec.} & \textbf{Rec.} & \textbf{F1} & \textbf{AUROC} \\
\midrule
Ghostbuster  & 0.638 & 0.606 & 0.622 & 0.667 \\
Glimpse  & 0.899 & 0.861 & 0.880 & 0.948 \\
\midrule
Binoculars  & 0.869 & 0.825 & 0.846 & 0.907 \\
FDG (falcon-7b)  & 0.670 & 0.635 & 0.652 & 0.708 \\
FDG (gpt-neo)  & 0.781 & 0.733 & 0.756 & 0.829 \\
Zippy (LZMA)  & 0.359 & 0.327 & 0.343 & 0.262 \\
\midrule
BiScope (Yelp)  & 0.706 & 0.691 & 0.699 & 0.726 \\
BiScope (Arxiv)  & 0.404 & 0.381 & 0.392 & 0.327 \\
BiScope (Essay)  & 0.841 & 0.761 & 0.799 & 0.805 \\
BiScope (Creative)  & 0.388 & 0.302 & 0.339 & 0.362 \\
DeTeCtive (MAGE)  & 0.470 & 0.127 & 0.200 & 0.477 \\
DeTeCtive (M4GT)  & 0.863 & 0.472 & 0.610 & 0.696 \\
DeTeCtive (TuringBench)  & 0.508 & 0.870 & 0.641 & 0.450 \\
RADAR  & 0.700 & 0.613 & 0.653 & 0.706 \\
Desklib  & 0.976 & 0.960 & 0.968 & 0.994 \\
E5-lora  & 0.417 & 0.361 & 0.387 & 0.362 \\
\bottomrule
    \end{tabular}
\caption{Benchmark model performance for the investigated systems on a balanced corpus of human-written and LLM-generated texts.}
\label{tab:benchmark}
\end{table}

We find that model performance is fairly disparate: the zero-shot models are mostly strong except for Zippy. The trained models depend heavily on the dataset: BiScope performance ranges in AUROC from 0.362 to 0.805, with the Essay variant performing best, likely because it best matches the evaluation data domain. These models tend to struggle when applied to new domains, and while this is important to note, our primary goal is not to compare or evaluate the exact performance of these models, but rather to assess whether the mistakes they are making significantly favor certain groups. For this reason, we proceed with our bias analysis using all models, with the understanding that some may be better or worse suited to this task.

We also aim for relative model independence to cover a broad range of potential systems. We calculate Pearson correlations between all models: only 5.5\% of pairs have correlation over 0.6, and none with correlation over 0.8, indicating models have weak to moderate correlation. The primary correlations are between BiScope models, where the Creative variant overlaps with the Essay and Yelp variants, and between the two FDG variants.\footnote{For more, see Appendix \ref{app:correlations}.}

    \begin{table*}[t!]
    \setlength\tabcolsep{1.2pt}

    \small
    \centering
\begin{tabular}{r|cccl|cccl|cccl|cccl}
    \toprule
    & \multicolumn{4}{c|}{\textbf{Gender}}    & \multicolumn{4}{c|}{\textbf{Race/Ethnicity}}     & \multicolumn{4}{c|}{\textbf{ELL Status}}     & \multicolumn{4}{c}{\textbf{Econ. Status}} \\
    \textbf{Model} & Diff. & Coef. & DA & & Diff. & Coef. & DA & & Diff. & Coef. & DA & & Diff. & Coef. & DA & \\ 
\midrule
Ghostbuster & \ovalcell[ForestGreen!10]{-.102} & \ovalcell[ForestGreen!2]{-.139$\ddag$} & \ovalcell[ForestGreen!4]{2.98} &  & \ovalcell[DarkOrchid!4]{.041} & \ovalcell[ForestGreen!4]{-.244$\ddag$} & 1.01 &  & \ovalcell[ForestGreen!14]{-.143} & \ovalcell[ForestGreen!18]{-.945$\ddag$} & \ovalcell[ForestGreen!9]{6.50} & \DisArrow & \ovalcell[ForestGreen!11]{-.120} & \ovalcell[ForestGreen!6]{-.338$\ddag$} & \ovalcell[ForestGreen!8]{5.87} & \DisArrow \\
Glimpse & \ovalcell[ForestGreen!4]{-.046} & \ovalcell[ForestGreen!6]{-.338$\ddag$} & 1.07 &  & \ovalcell[ForestGreen!1]{-.016} & \ovalcell[ForestGreen!20]{-.012$\ddag$} & 0.32 &  & \ovalcell[ForestGreen!4]{-.047} & \ovalcell[ForestGreen!23]{-.199$\ddag$} & 0.86 &  & \ovalcell[ForestGreen!3]{-.036} & \ovalcell[ForestGreen!8]{-.444$\ddag$} & 0.87 &  \\
Binoculars & \ovalcell[ForestGreen!4]{-.042} & \ovalcell[ForestGreen!3]{-.168$\ddag$} & 1.00 &  & \ovalcell[ForestGreen!0]{-.002} & \ovalcell[ForestGreen!10]{-.548$\ddag$} & 0.09 &  & \ovalcell[ForestGreen!1]{-.010} & \ovalcell[ForestGreen!9]{-.475$\ddag$} & 0.53 &  & \ovalcell[ForestGreen!5]{-.059} & \ovalcell[ForestGreen!5]{-.280$\ddag$} & \ovalcell[ForestGreen!3]{2.25} &  \\
FDG (falcon-7b) & \ovalcell[DarkOrchid!6]{.068} & \ovalcell[DarkOrchid!3]{.167$\ddag$} & \ovalcell[DarkOrchid!4]{2.80} &  & \ovalcell[DarkOrchid!0]{.007} & \ovalcell[DarkOrchid!7]{.361$\ddag$} & 0.14 &  & \ovalcell[DarkOrchid!0]{.003} & \ovalcell[DarkOrchid!3]{.195$\ddag$} & 1.13 &  & \ovalcell[DarkOrchid!7]{.080} & \ovalcell[DarkOrchid!3]{.185$\ddag$} & \ovalcell[DarkOrchid!6]{4.09} &  \\
FDG (gpt-neo) & \ovalcell[DarkOrchid!5]{.052} & \ovalcell[DarkOrchid!3]{.182$\ddag$} & \ovalcell[DarkOrchid!6]{4.65} &  & \ovalcell[DarkOrchid!1]{.018} & \ovalcell[DarkOrchid!5]{.267$\ddag$} & 0.85 &  & \ovalcell[ForestGreen!1]{-.019} & \ovalcell[ForestGreen!1]{-.081} & \ovalcell[ForestGreen!5]{3.87} &  & \ovalcell[DarkOrchid!4]{.047} & \ovalcell[DarkOrchid!2]{.115$\ddag$} & \ovalcell[DarkOrchid!5]{3.39} &  \\
Zippy (LZMA) & \ovalcell[ForestGreen!3]{-.036} & \ovalcell[ForestGreen!2]{-.146$\ddag$} & 1.39 &  & \ovalcell[DarkOrchid!0]{.009} & \ovalcell[DarkOrchid!9]{.471$\ddag$} & 0.26 &  & \ovalcell[ForestGreen!13]{-.135} & \ovalcell[ForestGreen!5]{-.286$\ddag$} & \ovalcell[ForestGreen!22]{14.84} & \DisArrow & \ovalcell[DarkOrchid!4]{.048} & \ovalcell[DarkOrchid!3]{.177$\ddag$} & 1.34 &  \\
\midrule
BiScope (Yelp) & \ovalcell[DarkOrchid!5]{.052} & \ovalcell[DarkOrchid!1]{.085} & \ovalcell[DarkOrchid!5]{3.36} &  & \ovalcell[ForestGreen!0]{-.001} & \ovalcell[DarkOrchid!2]{.141$\ddag$} & 0.01 &  & \ovalcell[DarkOrchid!2]{.022} & \ovalcell[DarkOrchid!1]{.090} & 0.52 &  & \ovalcell[DarkOrchid!7]{.074} & \ovalcell[DarkOrchid!4]{.224$\ddag$} & \ovalcell[DarkOrchid!15]{10.10} & \AdvArrow \\
BiScope (Arxiv) & \ovalcell[DarkOrchid!2]{.028} & \ovalcell[DarkOrchid!1]{.088$\ddag$} & 0.90 &  & \ovalcell[DarkOrchid!0]{.010} & \ovalcell[DarkOrchid!6]{.301$\ddag$} & 0.40 &  & -.035 & .042 & \ovalcell[DarkOrchid!7]{5.01} &  & \ovalcell[DarkOrchid!5]{.058} & \ovalcell[DarkOrchid!3]{.184$\ddag$} & \ovalcell[DarkOrchid!6]{4.06} &  \\
BiScope (Essay) & \ovalcell[DarkOrchid!1]{.020} & \ovalcell[DarkOrchid!1]{.082} & \ovalcell[DarkOrchid!4]{2.74} &  & .011 & .048 & 0.71 &  & \ovalcell[ForestGreen!1]{-.016} & \ovalcell[ForestGreen!5]{-.280$\ddag$} & \ovalcell[ForestGreen!7]{5.25} & \DisArrow & \ovalcell[DarkOrchid!2]{.024} & \ovalcell[DarkOrchid!3]{.166$\ddag$} & \ovalcell[DarkOrchid!9]{6.03} & \AdvArrow \\
BiScope (Creative) & \ovalcell[DarkOrchid!7]{.073} & \ovalcell[DarkOrchid!5]{.255$\ddag$} & \ovalcell[DarkOrchid!4]{2.95} &  & \ovalcell[ForestGreen!1]{-.016} & \ovalcell[DarkOrchid!6]{.311$\ddag$} & 0.14 &  & \ovalcell[ForestGreen!0]{-.009} & \ovalcell[DarkOrchid!3]{.162$\ddag$} & 1.81 &  & \ovalcell[DarkOrchid!10]{.102} & \ovalcell[DarkOrchid!6]{.344$\ddag$} & \ovalcell[DarkOrchid!9]{6.16} & \AdvArrow \\
DeTeCtive (MAGE) & \ovalcell[ForestGreen!5]{-.056} & \ovalcell[ForestGreen!3]{-.185$\ddag$} & \ovalcell[ForestGreen!8]{5.53} & \DisArrow & .008 & .038 & 0.18 &  & \ovalcell[ForestGreen!18]{-.180} & \ovalcell[ForestGreen!24]{-.249$\ddag$} & \ovalcell[ForestGreen!50]{79.12} & \DisArrow & \ovalcell[ForestGreen!0]{-.004} & \ovalcell[DarkOrchid!2]{.105} & 0.54 &  \\
DeTeCtive (M4GT) & .008 & -.058 & 0.25 &  & \ovalcell[ForestGreen!0]{-.008} & \ovalcell[ForestGreen!2]{-.109} & 1.49 &  & .009 & .083 & 0.26 &  & \ovalcell[DarkOrchid!1]{.017} & \ovalcell[DarkOrchid!5]{.288$\ddag$} & \ovalcell[DarkOrchid!6]{4.39} &  \\
DeTeCtive (TuringBench) & .004 & .010 & 0.15 &  & \ovalcell[ForestGreen!0]{-.004} & \ovalcell[DarkOrchid!4]{.238$\ddag$} & 0.13 &  & \ovalcell[ForestGreen!0]{-.008} & \ovalcell[ForestGreen!6]{-.344$\ddag$} & 1.96 &  & .014 & .129 & 1.31 &  \\
RADAR & .007 & .037 & 0.70 &  & \ovalcell[DarkOrchid!4]{.049} & \ovalcell[ForestGreen!2]{-.111$\ddag$} & \ovalcell[ForestGreen!9]{6.48} & \DisArrow & \ovalcell[DarkOrchid!0]{.006} & \ovalcell[ForestGreen!9]{-.470$\ddag$} & 0.84 &  & \ovalcell[ForestGreen!3]{-.035} & \ovalcell[ForestGreen!6]{-.310$\ddag$} & \ovalcell[ForestGreen!5]{3.72} &  \\
Desklib & -.002 & .001 & 0.04 &  & \ovalcell[ForestGreen!0]{-.002} & \ovalcell[ForestGreen!7]{-.396$\ddag$} & 1.02 &  & \ovalcell[ForestGreen!0]{-.001} & \ovalcell[ForestGreen!5]{-.274$\ddag$} & 0.04 &  & -.007 & .027 & 0.73 &  \\
E5-lora & \ovalcell[ForestGreen!4]{-.044} & \ovalcell[DarkOrchid!1]{.074} & 0.37 &  & \ovalcell[DarkOrchid!10]{.103} & \ovalcell[DarkOrchid!3]{.186$\ddag$} & \ovalcell[DarkOrchid!7]{5.15} & \AdvArrow & \ovalcell[ForestGreen!2]{-.029} & \ovalcell[ForestGreen!4]{-.212$\ddag$} & 0.21 &  & \ovalcell[ForestGreen!12]{-.125} & \ovalcell[ForestGreen!5]{-.257$\ddag$} & \ovalcell[ForestGreen!8]{5.49} & \DisArrow \\
\bottomrule
\end{tabular}
\caption{Model performance differences (Diff.), attribute coeffecients (Coef., $p < 1.56e-4$) and Dominance Analysis scores (DA) for each model. \AdvArrow{} indicates the advantaged class is more likely classified as machine-generated; \DisArrow{} indicates the disadvantaged class is more likely machine-generated.}
\label{tab:significance}
\end{table*}

\section{Logistic Regression Analysis}
    \label{sec:lr}
    To study bias, we need a methodology that can account for the relationships and confounds present in the data. The attributes we study are unlikely to be independent, and additional factors may influence model performance. To handle this, we perform our analysis based on the methodology of \newcite{dayanik-2022}, who outline procedures for identifying bias in natural language processing systems where there may be many factors involved.
    
    We train a logistic regression model over relevant features as well as confounds to predict the error of the model, and use feature coefficients as well as dominance analysis to assess the impact of each attribute. For bias variables, we use the sensitive attributes: gender (male/female), race/ethnicity (White/non-White), English language-learner status (no/yes), and economic status (not disadvantaged/disadvantaged). We then define covariates, which are other potential predictors of error: we use perplexity via the \texttt{opt-iml-1.3b} model \cite{iyer-2023} and length of the text in words (defined by splitting on whitespace). We calculate variance inflation factor (VIF) for each bias variable and covariate, finding the values to all be below 4, therefore suitable with minimal multicollinearity. We report the following:

    \paragraph{Performance Difference (Diff.):} The difference in model performance for each attribute. This is calculated as the mean score for the advantaged class minus that of the disadvantaged class. Higher scores indicate that the advantaged class is more likely to be classified as machine-generated; lower scores indicate the disadvantaged class is more likely to be classified as machine-generated.
    \paragraph{Coefficient (Coef.):} The coefficient for the given attribute in the regression model. Positive scores indicate the advantaged class predicts higher machine-generated scores; negative scores indicate the disadvantaged class. For all experiments, we use a significance threshold of 0.01 with Bonferroni correction, using the number of models (16) and the number of categories, yielding a $p$ value of $0.01 / (categories \times models)$; we note the specific values in each experiment.  
    \paragraph{Dominance analysis (DA):} This indicates the strength of this attribute's contribution in the logistic regression model. We report dominance as percentage of relative importance (e.g., 4.73 indicates that 4.73\% of the prediction comes from this attribute). We consider dominance scores over 5\% to be meaningful.
    
    \paragraph{} Significance tests are useful for detecting the presence of systematic differences, but do not capture the magnitude of difference \cite{dayanik-2022,stanczak-2021}. Hence, we are most interested in cases where both the coefficient from the model is significant (indicating a significant relationship between the attribute and model performance) and the dominance factor is strong ($> 5$, indicating that at least 5\% of the difference in performance is due to this attribute): these cases are marked with \AdvArrow{} (the advantaged class is more likely to be classified as machine-generated) and \DisArrow{} (the disadvantaged class is more likely to be classified as machine-generated). Table \ref{tab:significance} shows these results over all models and categories.

\subsection{General Analysis}
We observe relatively few instances where model coefficients are significant and dominance analysis indicates a strong influence of the corresponding attribute (12 out of 64 total observations). These cases also tend to be inconsistent, showing no systematic preference for either advantaged or disadvantaged groups. The primary exception is ELL status, where most models incorrectly classify essays written by ELL students as machine-generated; four models exhibit both significant coefficients and dominance analysis results. We further analyze differences by model and attribute.

\subsection{Model Analysis}
Most models exhibit inconsistent and minor biases across all categories. The two GPT-based models (Ghostbuster and Glimpse), as well as Binoculars, RADAR, and the DeTeCtive (MAGE) variant, tend to misclassify the disadvantaged population as machine-generated, although the associated dominance is often minimal: while the effects are statistically significant, the attribute plays only a minor role in the resulting classification. FDG models are relatively inconsistent, while most trained models tend to misclassify essays from both ELL students and those without economic disadvantages as machine-generated.

\subsection{Attribute Analysis}
We observe relatively limited impact from gender and race in this analysis, though subgroup analysis may provide more insight. While many models exhibit significant coefficients, these are not reflected in the dominance analysis, suggesting a limited role in classification decisions. The affected groups also vary: different models and variants exhibit minor biases in both directions.

ELL status shows consistent negative effects: ELL essays are more likely to be classified as machine-generated by most models. While this general trend aligns with prior research indicating that ELL students are unfairly treated by detection systems, we note that the magnitude of this effect is typically small.

Economic status shows a relatively strong but mixed effect. Essays from students without economic disadvantage are more likely to be classified as machine-generated by BiScope and FDG models. However, other zero-shot and fine-tuned models present potential risks: the GPT-based Ghostbuster and Glimpse models, as well as all three trained variants, exhibit negative coefficients, which in two cases correspond with higher dominance values.

    \begin{table*}[]
    \setlength\tabcolsep{1.5pt}

    \small
    \centering
\begin{tabularx}{\textwidth}
{r | Y Y | Y Y | Y Y | Y Y | Y Y | Y Y | Y Y | Y Y }
\toprule
Gender & \multicolumn{8}{c|}{Male} & \multicolumn{8}{c}{Female} \\
Race & \multicolumn{4}{c|}{White} & \multicolumn{4}{c|}{Non-White} & \multicolumn{4}{c|}{White} & \multicolumn{4}{c}{Non-White} \\
ELL & \multicolumn{2}{c|}{No} & \multicolumn{2}{c|}{Yes} & \multicolumn{2}{c|}{No} & \multicolumn{2}{c|}{Yes} & \multicolumn{2}{c|}{No} & \multicolumn{2}{c|}{Yes} & \multicolumn{2}{c|}{No} & \multicolumn{2}{c}{Yes} \\
Econ. Disadvantage & \multicolumn{1}{c|}{No} & \multicolumn{1}{c|}{Yes} & \multicolumn{1}{c|}{No} & \multicolumn{1}{c|}{Yes} & \multicolumn{1}{c|}{No} & \multicolumn{1}{c|}{Yes} & \multicolumn{1}{c|}{No} & \multicolumn{1}{c|}{Yes} & \multicolumn{1}{c|}{No} & \multicolumn{1}{c|}{Yes} & \multicolumn{1}{c|}{No} & \multicolumn{1}{c|}{Yes} & \multicolumn{1}{c|}{No} & \multicolumn{1}{c|}{Yes} & \multicolumn{1}{c|}{No} & Yes \\
\midrule
Ghostbuster & \cellcolor{gray!25} \ovalcell[ForestGreen!25]{-.09} & \cellcolor{gray!25} {--} & \cellcolor{White} {--} & \cellcolor{White} {--} & \cellcolor{gray!25} \ovalcell[ForestGreen!25]{-.06} & \cellcolor{gray!25} \ovalcell[DarkOrchid!25]{.06} & \cellcolor{White} \ovalcell[DarkOrchid!25]{.10} & \cellcolor{White} \ovalcell[DarkOrchid!25]{.15} & \cellcolor{gray!25} \ovalcell[ForestGreen!25]{-.16} & \cellcolor{gray!25} {--} & \cellcolor{White} {--} & \cellcolor{White} {--} & \cellcolor{gray!25} \ovalcell[ForestGreen!25]{-.11} & \cellcolor{gray!25} {--} & \cellcolor{White} \ovalcell[DarkOrchid!25]{.12} & \cellcolor{White} \ovalcell[DarkOrchid!25]{.13}\\
Glimpse & \cellcolor{gray!25} \ovalcell[ForestGreen!25]{-.05} & \cellcolor{gray!25} \ovalcell[ForestGreen!25]{-.03} & \cellcolor{White} {--} & \cellcolor{White} {--} & \cellcolor{gray!25} {--} & \cellcolor{gray!25} {--} & \cellcolor{White} {--} & \cellcolor{White} \ovalcell[DarkOrchid!25]{.03} & \cellcolor{gray!25} \ovalcell[ForestGreen!25]{-.05} & \cellcolor{gray!25} {--} & \cellcolor{White} {--} & \cellcolor{White} {--} & \cellcolor{gray!25} {--} & \cellcolor{gray!25} \ovalcell[DarkOrchid!25]{.02} & \cellcolor{White} \ovalcell[DarkOrchid!25]{.05} & \cellcolor{White} \ovalcell[DarkOrchid!25]{.07}\\
Binoculars & \cellcolor{gray!25} \ovalcell[DarkOrchid!25]{.05} & \cellcolor{gray!25} {--} & \cellcolor{White} {--} & \cellcolor{White} {--} & \cellcolor{gray!25} {--} & \cellcolor{gray!25} \ovalcell[ForestGreen!25]{-.04} & \cellcolor{White} {--} & \cellcolor{White} \ovalcell[ForestGreen!25]{-.03} & \cellcolor{gray!25} \ovalcell[DarkOrchid!25]{.05} & \cellcolor{gray!25} {--} & \cellcolor{White} {--} & \cellcolor{White} {--} & \cellcolor{gray!25} {--} & \cellcolor{gray!25} \ovalcell[ForestGreen!25]{-.04} & \cellcolor{White} {--} & \cellcolor{White} {--}\\
FDG (falcon-7b) & \cellcolor{gray!25} \ovalcell[DarkOrchid!25]{.08} & \cellcolor{gray!25} {--} & \cellcolor{White} {--} & \cellcolor{White} {--} & \cellcolor{gray!25} {--} & \cellcolor{gray!25} \ovalcell[ForestGreen!25]{-.07} & \cellcolor{White} {--} & \cellcolor{White} {--} & \cellcolor{gray!25} \ovalcell[DarkOrchid!25]{.08} & \cellcolor{gray!25} {--} & \cellcolor{White} {--} & \cellcolor{White} {--} & \cellcolor{gray!25} {--} & \cellcolor{gray!25} \ovalcell[ForestGreen!25]{-.07} & \cellcolor{White} {--} & \cellcolor{White} {--}\\
FDG (gpt-neo) & \cellcolor{gray!25} \ovalcell[DarkOrchid!25]{.06} & \cellcolor{gray!25} {--} & \cellcolor{White} {--} & \cellcolor{White} {--} & \cellcolor{gray!25} {--} & \cellcolor{gray!25} \ovalcell[ForestGreen!25]{-.05} & \cellcolor{White} {--} & \cellcolor{White} {--} & \cellcolor{gray!25} \ovalcell[DarkOrchid!25]{.05} & \cellcolor{gray!25} {--} & \cellcolor{White} {--} & \cellcolor{White} {--} & \cellcolor{gray!25} {--} & \cellcolor{gray!25} \ovalcell[ForestGreen!25]{-.06} & \cellcolor{White} {--} & \cellcolor{White} {--}\\
Zippy (LZMA) & \cellcolor{gray!25} {--} & \cellcolor{gray!25} \ovalcell[ForestGreen!25]{-.11} & \cellcolor{White} {--} & \cellcolor{White} {--} & \cellcolor{gray!25} {--} & \cellcolor{gray!25} \ovalcell[ForestGreen!25]{-.07} & \cellcolor{White} \ovalcell[DarkOrchid!25]{.14} & \cellcolor{White} \ovalcell[DarkOrchid!25]{.11} & \cellcolor{gray!25} {--} & \cellcolor{gray!25} \ovalcell[ForestGreen!25]{-.10} & \cellcolor{White} {--} & \cellcolor{White} {--} & \cellcolor{gray!25} {--} & \cellcolor{gray!25} \ovalcell[ForestGreen!25]{-.08} & \cellcolor{White} \ovalcell[DarkOrchid!25]{.14} & \cellcolor{White} \ovalcell[DarkOrchid!25]{.12}\\
BiScope (Yelp) & \cellcolor{gray!25} \ovalcell[DarkOrchid!25]{.07} & \cellcolor{gray!25} {--} & \cellcolor{White} {--} & \cellcolor{White} {--} & \cellcolor{gray!25} {--} & \cellcolor{gray!25} \ovalcell[ForestGreen!25]{-.05} & \cellcolor{White} {--} & \cellcolor{White} {--} & \cellcolor{gray!25} \ovalcell[DarkOrchid!25]{.10} & \cellcolor{gray!25} {--} & \cellcolor{White} {--} & \cellcolor{White} {--} & \cellcolor{gray!25} {--} & \cellcolor{gray!25} \ovalcell[ForestGreen!25]{-.06} & \cellcolor{White} {--} & \cellcolor{White} {--}\\
BiScope (Arxiv) & \cellcolor{gray!25} \ovalcell[DarkOrchid!25]{.05} & \cellcolor{gray!25} \ovalcell[ForestGreen!25]{-.04} & \cellcolor{White} {--} & \cellcolor{White} {--} & \cellcolor{gray!25} {--} & \cellcolor{gray!25} \ovalcell[ForestGreen!25]{-.05} & \cellcolor{White} {--} & \cellcolor{White} \ovalcell[DarkOrchid!25]{.04} & \cellcolor{gray!25} \ovalcell[DarkOrchid!25]{.06} & \cellcolor{gray!25} {--} & \cellcolor{White} {--} & \cellcolor{White} {--} & \cellcolor{gray!25} {--} & \cellcolor{gray!25} \ovalcell[ForestGreen!25]{-.07} & \cellcolor{White} {--} & \cellcolor{White} {--}\\
BiScope (Essay) & \cellcolor{gray!25} \ovalcell[DarkOrchid!25]{.04} & \cellcolor{gray!25} {--} & \cellcolor{White} {--} & \cellcolor{White} {--} & \cellcolor{gray!25} {--} & \cellcolor{gray!25} {--} & \cellcolor{White} {--} & \cellcolor{White} {--} & \cellcolor{gray!25} \ovalcell[DarkOrchid!25]{.04} & \cellcolor{gray!25} {--} & \cellcolor{White} {--} & \cellcolor{White} {--} & \cellcolor{gray!25} {--} & \cellcolor{gray!25} \ovalcell[ForestGreen!25]{-.04} & \cellcolor{White} {--} & \cellcolor{White} {--}\\
BiScope (Creative) & \cellcolor{gray!25} \ovalcell[DarkOrchid!25]{.08} & \cellcolor{gray!25} \ovalcell[ForestGreen!25]{-.04} & \cellcolor{White} {--} & \cellcolor{White} {--} & \cellcolor{gray!25} {--} & \cellcolor{gray!25} \ovalcell[ForestGreen!25]{-.10} & \cellcolor{White} {--} & \cellcolor{White} {--} & \cellcolor{gray!25} \ovalcell[DarkOrchid!25]{.13} & \cellcolor{gray!25} {--} & \cellcolor{White} {--} & \cellcolor{White} {--} & \cellcolor{gray!25} \ovalcell[DarkOrchid!25]{.04} & \cellcolor{gray!25} \ovalcell[ForestGreen!25]{-.10} & \cellcolor{White} {--} & \cellcolor{White} {--}\\
DeTeCtive (MAGE) & \cellcolor{gray!25} \ovalcell[ForestGreen!25]{-.03} & \cellcolor{gray!25} \ovalcell[ForestGreen!25]{-.08} & \cellcolor{White} {--} & \cellcolor{White} \ovalcell[DarkOrchid!25]{.10} & \cellcolor{gray!25} \ovalcell[ForestGreen!25]{-.04} & \cellcolor{gray!25} \ovalcell[ForestGreen!25]{-.06} & \cellcolor{White} \ovalcell[DarkOrchid!25]{.15} & \cellcolor{White} \ovalcell[DarkOrchid!25]{.16} & \cellcolor{gray!25} {--} & \cellcolor{gray!25} \ovalcell[ForestGreen!25]{-.08} & \cellcolor{White} {--} & \cellcolor{White} {--} & \cellcolor{gray!25} \ovalcell[ForestGreen!25]{-.03} & \cellcolor{gray!25} \ovalcell[ForestGreen!25]{-.07} & \cellcolor{White} \ovalcell[DarkOrchid!25]{.15} & \cellcolor{White} \ovalcell[DarkOrchid!25]{.16}\\
DeTeCtive (M4GT) & \cellcolor{gray!25} {--} & \cellcolor{gray!25} {--} & \cellcolor{White} {--} & \cellcolor{White} {--} & \cellcolor{gray!25} {--} & \cellcolor{gray!25} {--} & \cellcolor{White} {--} & \cellcolor{White} {--} & \cellcolor{gray!25} {--} & \cellcolor{gray!25} {--} & \cellcolor{White} {--} & \cellcolor{White} {--} & \cellcolor{gray!25} {--} & \cellcolor{gray!25} {--} & \cellcolor{White} {--} & \cellcolor{White} {--}\\
DeTeCtive (TuringBench) & \cellcolor{gray!25} {--} & \cellcolor{gray!25} {--} & \cellcolor{White} {--} & \cellcolor{White} {--} & \cellcolor{gray!25} {--} & \cellcolor{gray!25} \ovalcell[ForestGreen!25]{-.02} & \cellcolor{White} {--} & \cellcolor{White} {--} & \cellcolor{gray!25} \ovalcell[DarkOrchid!25]{.01} & \cellcolor{gray!25} {--} & \cellcolor{White} {--} & \cellcolor{White} {--} & \cellcolor{gray!25} {--} & \cellcolor{gray!25} \ovalcell[ForestGreen!25]{-.01} & \cellcolor{White} {--} & \cellcolor{White} {--}\\
RADAR & \cellcolor{gray!25} {--} & \cellcolor{gray!25} \ovalcell[ForestGreen!25]{-.08} & \cellcolor{White} {--} & \cellcolor{White} {--} & \cellcolor{gray!25} {--} & \cellcolor{gray!25} \ovalcell[ForestGreen!25]{-.04} & \cellcolor{White} {--} & \cellcolor{White} {--} & \cellcolor{gray!25} \ovalcell[DarkOrchid!25]{.07} & \cellcolor{gray!25} {--} & \cellcolor{White} {--} & \cellcolor{White} {--} & \cellcolor{gray!25} \ovalcell[DarkOrchid!25]{.05} & \cellcolor{gray!25} {--} & \cellcolor{White} {--} & \cellcolor{White} {--}\\
Desklib & \cellcolor{gray!25} {--} & \cellcolor{gray!25} {--} & \cellcolor{White} {--} & \cellcolor{White} {--} & \cellcolor{gray!25} {--} & \cellcolor{gray!25} {--} & \cellcolor{White} {--} & \cellcolor{White} {--} & \cellcolor{gray!25} {--} & \cellcolor{gray!25} {--} & \cellcolor{White} {--} & \cellcolor{White} {--} & \cellcolor{gray!25} {--} & \cellcolor{gray!25} {--} & \cellcolor{White} {--} & \cellcolor{White} {--}\\
E5-lora & \cellcolor{gray!25} \ovalcell[ForestGreen!25]{-.03} & \cellcolor{gray!25} \ovalcell[DarkOrchid!25]{.12} & \cellcolor{White} {--} & \cellcolor{White} {--} & \cellcolor{gray!25} {--} & \cellcolor{gray!25} \ovalcell[DarkOrchid!25]{.14} & \cellcolor{White} \ovalcell[DarkOrchid!25]{.06} & \cellcolor{White} \ovalcell[DarkOrchid!25]{.08} & \cellcolor{gray!25} \ovalcell[ForestGreen!25]{-.18} & \cellcolor{gray!25} {--} & \cellcolor{White} \ovalcell[ForestGreen!25]{-.14} & \cellcolor{White} {--} & \cellcolor{gray!25} \ovalcell[ForestGreen!25]{-.16} & \cellcolor{gray!25} {--} & \cellcolor{White} {--} & \cellcolor{White} {--}\\
\bottomrule
\end{tabularx}
\belowcaptionskip=10pt
\caption{Results for subgroup analysis: we report the differences in error from the given subgroup to those not in that subgroup, with \ovalcell[DarkOrchid!25]{positive} scores indicating the subgroup is more likely classified as machine generated and \ovalcell[ForestGreen!25]{negative} indicating the opposite. We report only statistically significant differences ($p < 3.91e-5$).}
\label{tab:subgroup}
\end{table*}

\subsection{Overall Results}
    From the regression analysis, we conclude the following: (1) ELL status appears to be a major contributing factor, with ELL student essays more likely to be classified as machine-generated, consistent with prior research; (2) economic status is an important variable, although results vary across model types; and (3) there is relatively little evidence of bias related to race or gender.

\begin{figure}[t!]
\includegraphics[width=.49\textwidth]{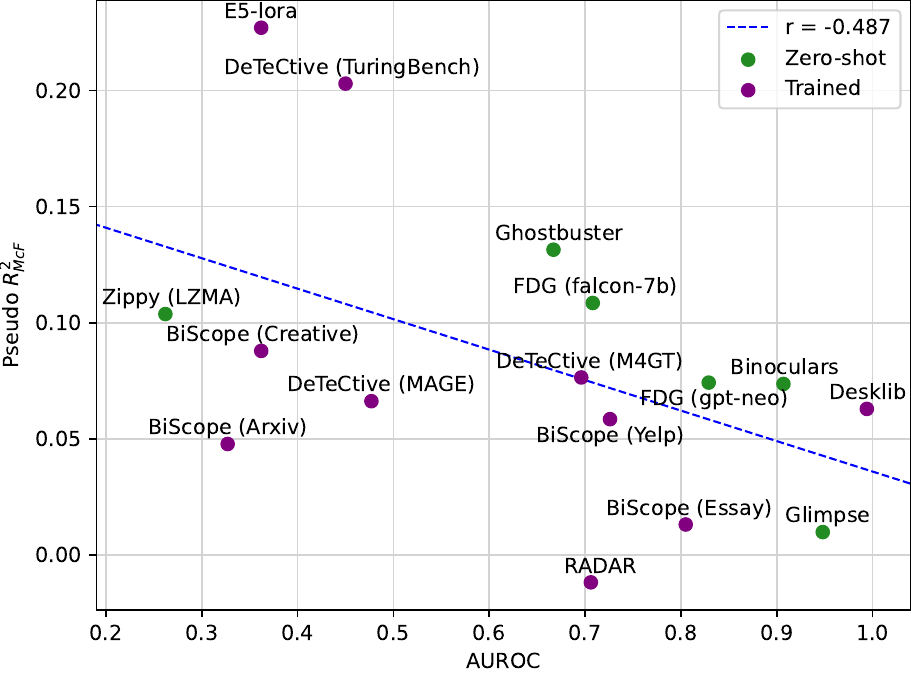}
\caption{Pseudo-$R^2$ values from the regression analysis plotted against AUROC scores for each model.}
\label{fig:biascorr}
\end{figure}

We further examine model performance in relation to overall bias. In Figure \ref{fig:biascorr}, we plot each model’s AUROC score against its McFadden’s pseudo-$R^2$ value from the regression analysis. While $R^2$ typically measures variance explained in linear regression, this approximation is used here as a general indicator of bias in logistic regression models. We observe a general negative correlation: as model performance (AUROC) increases, the estimated bias ($R^2_{McF}$) decreases ($r = -0.486$). This trend suggests that higher-performing models may exhibit lower bias and thus reduce potential harms.

\section{Subgroup Analysis}
While regression analysis offers a broad overview of biases by attribute, it may obscure subgroup-specific effects. To explore these, we partition the dataset into 16 subgroups representing all combinations of the four attributes under study. We then perform pairwise z-tests comparing each subgroup’s scores with those of the remaining dataset, identifying statistically significant differences in classification. Results are presented in Table \ref{tab:subgroup}.

This analysis reveals several notable findings not evident in the overall regression results. Although ELL essays are generally more likely to be classified as machine-generated by a large number of models, this effect is disproportionately concentrated among non-White students. Specifically, non-White ELL essays ($n = 9,443$) are more frequently misclassified by seven different models, compared to only one model for their White counterparts ($n = 1,221$). Moreover, this effect is more pronounced among males: three additional models exhibit significant differences for male non-White ELL essays compared to females.

   \begin{table}[t!]
    \setlength\tabcolsep{1.pt}
    \small
    \centering
\begin{tabular}
{r|cccccc}
\textbf{Model} & \rotatebox{90}{\textbf{W}} & \rotatebox{90}{\textbf{H/L}} & \rotatebox{90}{\textbf{A/PI}} & \rotatebox{90}{\textbf{B/AA}} & \rotatebox{90}{\textbf{Two+}} & \rotatebox{90}{\textbf{AI/AN}} \\ 
\midrule
Ghostbuster & -- & -- & -- & -- & -- & -- \\
Glimpse & \ovalcell[ForestGreen!25]{-.05} & \ovalcell[DarkOrchid!25]{.03} & -- & \ovalcell[DarkOrchid!25]{.05} & -.02 & -.02 \\
Binoculars & -- & -- & -- & -- & -- & -- \\
FDG (falcon-7b) & \ovalcell[DarkOrchid!25]{.04} & \ovalcell[ForestGreen!25]{-.02} & \ovalcell[DarkOrchid!25]{.06} & \ovalcell[ForestGreen!25]{-.07} & -.01 & .04 \\
FDG (gpt-neo) & \ovalcell[DarkOrchid!25]{.04} & -.01 & .02 & \ovalcell[ForestGreen!25]{-.06} & -.02 & .05 \\
Zippy (LZMA) & \ovalcell[ForestGreen!25]{-.02} & \ovalcell[DarkOrchid!25]{.05} & \ovalcell[DarkOrchid!25]{.10} & \ovalcell[ForestGreen!25]{-.07} & \ovalcell[ForestGreen!25]{-.07} & -.04 \\
\midrule
BiScope (Yelp) & -- & \ovalcell[ForestGreen!25]{-.00} & -- & -- & -- & -- \\
BiScope (Arxiv) & \ovalcell[DarkOrchid!25]{.02} & -.01 & \ovalcell[DarkOrchid!25]{.02} & \ovalcell[ForestGreen!25]{-.03} & -.01 & .01 \\
BiScope (Essay) & -- & -- & -- & -- & -- & -- \\
BiScope (Creative) & -- & -- & -- & -- & -- & -- \\
DeTeCtive (MAGE) & \ovalcell[ForestGreen!25]{-.06} & \ovalcell[DarkOrchid!25]{.08} & \ovalcell[DarkOrchid!25]{.05} & \ovalcell[ForestGreen!25]{-.04} & \ovalcell[ForestGreen!25]{-.06} & .02 \\
DeTeCtive (M4GT) & .01 & -.01 & .02 & -.01 & .01 & -.01 \\
DeTeCtive (TuringBench) & -- & .01 & .01 & \ovalcell[ForestGreen!25]{-.02} & -- & .01 \\
RADAR & -- & -- & \ovalcell[DarkOrchid!25]{.05} & -.01 & -.02 & -.01 \\
Desklib & -- & -- & -.01 & -- & .01 & .04 \\
E5-lora & \ovalcell[ForestGreen!25]{-.05} & \ovalcell[DarkOrchid!25]{.04} & \ovalcell[ForestGreen!25]{-.09} & \ovalcell[DarkOrchid!25]{.08} & -.03 & -.01 \\
\bottomrule
\end{tabular}
\caption{Performance differences on race/ethnicity groups. Highlighted values indicate significance based on z-scores between this attribute and the rest of the dataset ($p < 1.04e-4$).}
\label{tab:race}
\end{table}

We therefore need to revise our earlier assessment that bias related to race and gender is minimal. Subgroup analysis suggests that both race and gender play a substantial role, highlighting the need for more rigorous intersectional analysis when approaching potential biases.

For non-ELL essays, some differences are significant, but results are inconsistent. Interestingly, essays from ELL students with economic disadvantages are often less likely to be misclassified, though this outcome varies considerably by model.

\section{Race/Ethnicity}
\label{sec:race}

We initially conducted analysis using a simplified binary race/ethnicity classification (White vs. non-White). Here, we extend our analysis to explore individual race and ethnicity categories. The dataset includes six groups: White (W), Hispanic/Latino (H/L), Asian/Pacific Islander (A/PI), Black/African American (B/AA), Two or More Races/Other (Two+), and American Indian/Alaskan Native (AI/AN). We evaluate model performance on each individual group and compare it against the full dataset in Table \ref{tab:race}.

First, we note that no models exhibited significant performance differences for the AI/AN subgroup, likely due to the small sample size ($n = 208$). This lack of significance should not be interpreted as conclusive evidence of no bias, but rather as an indication that the dataset is underpowered for detecting such effects. Further investigation with more representative data is warranted.

Two groups, however, show consistent disparities: A/PI essays are more likely to be classified as machine-generated by most models, while B/AA essays are less likely to be misclassified in this way. Results for H/L and W essays are inconsistent, while the Two+ category shows significant negative effects in only two models.

\section{Human Performance}
Identifying machine-generated text remains a challenging task for humans \cite{dugan-2023,ethayarajh-2022,clark-2021}. \newcite{lee-2025} finds that, even with the aid of collaborative tools, human accuracy in this task reaches only 57\%. We evaluate human performance with respect to potential bias: given the same dataset, do human annotators exhibit biases comparable to those observed in automated detection systems?

To investigate this, we selected a balanced subsample of our corpus consisting of 800 total texts, with at least 100 examples from each group across the four key attributes: gender (male/female), race (White/non-White), ELL status (ELL/non-ELL), and economic status (disadvantaged/not disadvantaged). For each text, we used Claude Sonnet 3.5 \cite{claude3.5} to generate a continuation based on the first twenty tokens.\footnote{See Appendix \ref{app:prompts} for full prompt details.} This process resulted in a dataset containing equal numbers of human-written and LLM-generated texts, balanced across all demographic categories.

Three expert annotators were each assigned a subset of these texts ($231 < n < 318$), including a minimum of 25 human-written samples for each sensitive attribute. Annotators were instructed to classify each text as either human-written or machine-generated. We then applied the same logistic regression analysis used in Section \ref{sec:lr} to assess potential biases in human predictions. Performance differences are presented in Table \ref{tab:human}.\footnote{Full annotator details are provided in Appendix \ref{app:annotators}.}

\begin{table}[t!]
    \centering
    \small
\begin{tabular}{l|c|cccc}
\textbf{Annotator} & \rotatebox{90}{\textbf{Accuracy}} & \rotatebox{90}{\textbf{Gender}} & \rotatebox{90}{\textbf{Race/Ethnicity}} & \rotatebox{90}{\textbf{ELL Status}} & \rotatebox{90}{\textbf{Econ. Status}} \\
\midrule
Ann. 1 & 0.492 & 0.060 & -0.069 & 0.013 & 0.150 \\
Ann. 2 & 0.449 & -0.096 & -0.033 & 0.131 & 0.011 \\
Ann. 3 & 0.526 & -0.011 & -0.004 & 0.018 & 0.183 \\
\bottomrule
\end{tabular}
\caption{Differences in human classifications based on attributes. Positive numbers indicate the advantaged attribute is more likely classified as machine-generated; negative indicate the opposite. No results were indicated as significant for $p < .01$.}
\label{tab:human}
\end{table}

Our results align with previous findings regarding human performance: annotators performed at approximately chance. However, we found no significant differences in classification based on the studied attributes. While slightly elevated coefficients were observed for economic status (mirroring trends in the system evaluations), these were not statistically significant.

\section{Conclusions}
This work investigates bias in machine-generated text detection systems across four key attributes: gender, race/ethnicity, ELL status, and economic status. We find that several models tend to disproportionately affect disadvantaged groups; essays written by ELL students are more frequently misclassified as machine-generated, and this effect is particularly pronounced among non-White students. We also observe that while human annotators perform poorly at this task, they do not exhibit significant biases.

The key takeaway for practitioners is the critical importance of understanding the behavior and limitations of machine-generated text detection models. Misclassification presents a substantial risk, not only in this context but in other domains where such models may be applied. Our findings show no singular or consistent bias across all systems, underscoring the need for case-by-case evaluation. To ensure fairness, models and their predictions must be carefully scrutinized for disproportionate impacts on disadvantaged populations. AI developers and regulators can support this goal by creating and adopting datasets and metrics such as those proposed here that allow for the detection and mitigation of bias before real-world deployment.

\section{Limitations}
This study represents an initial step in analyzing bias in machine-generated text detection systems, but it is necessarily constrained in several ways.

\subsection{Models}
We examine only a limited subset of models commonly used for detecting machine-generated text. Our selection criteria emphasized public availability, broad use, and general applicability. While we aimed for methodological diversity, many relevant models remain outside the scope of this analysis, and our findings should not be assumed to generalize across all possible systems.

\subsection{Dataset}
Our evaluation data is similarly constrained. It consists entirely of student essay writing, drawn from three datasets produced by the same organization. All datasets are in English, and written by students in the United States. This choice was driven by three considerations: (1) the datasets are publicly available and include detailed demographic information (an uncommon feature); (2) we had access to a corresponding machine-generated dataset (\textsc{outfox}), facilitating benchmarking; and (3) the education domain represents a high-stakes use case, where misclassification could cause significant harm.

The consequence of these advantages is limited generalizability. Our findings may not extend to other text domains, and the dataset reflects a narrow slice of the broader population. Accordingly, the biases observed here may not reflect those that would occur when systems are evaluated on other demographics or styles of text.

\subsection{Categories of Gender and Race}
We acknowledge concerns regarding binary gender labels and predefined racial categories, as addressed in Section \ref{sec:data}, and this remains a limitation. Prior work warns that such categorization may reinforce essentialist or harmful views of identity. We are constrained here by the demographic labels provided in the datasets. Future research should explore more inclusive and representative identity categorizations.

\subsection{Statistical Methods}
Numerous statistical approaches exist for evaluating model fairness, and this remains an active area of research in bias in NLP. We sought to minimize methodological inconsistencies by following the statistical framework proposed by \newcite{dayanik-2022}, whose work closely aligns with our use case. However, we recognize that alternative methodologies could yield different insights.

\subsection{Human Annotation}
Our human annotation effort involved only three expert annotators. This limited scope was a function of prioritizing expertise over general human judgment. Although preliminary, our results suggest that human evaluators may introduce less bias than automated systems. However, the small sample size restricts the generalizability of this finding.

Overall, we acknowledge that this study is constrained by time, resources, and dataset availability. Many of the questions raised here warrant further investigation at larger scales and across more diverse settings.

\bibliography{acl_latex}

\appendix
\section{Data}
\label{app:data}

Statistics for the human-written dataset that was curated for our experiments are shown in Table \ref{tab:dataset}.

\begin{table*}[t!]
\small
\begin{tabular}{lc|cc|cc|cc|cc}
    \toprule
    & & \multicolumn{2}{c|}{\textbf{Gender}} &  \multicolumn{2}{c|}{\textbf{Race/Ethnicity}} &  \multicolumn{2}{c|}{\textbf{ELL}} &  \multicolumn{2}{c}{\textbf{Economic Disadvantage}} \\
    & Total & Male & Female & White & Nonwhite & No & Yes & No & Yes \\
    \midrule
   \textsc{persuade v2.0} & 24695 & 12074 & 12621 & 11282 & 13413 & 22451 & 2244 & 11003 & 13692\\
    \textsc{asap v2.0} & 24728 & 12498 & 12230 & 9841 & 14887 & 20991 & 3737 & 7933 & 16795\\
    \textsc{ellipse} & 6482 & 3636 & 2846 & 471 & 6011 & 0 & 6482 & 1974 & 4508\\
    \midrule
    Combined (clean) & 41743 & 21277 & 20466 & 15078 & 26665 & 31079 & 10664 & 18188 & 23555 \\
    \bottomrule
    \end{tabular}
\caption{Counts for each attribute in our combined corpus. Note that the \textsc{ellipse} corpus is designed to capture ELL speakers, and thus contains only that group.}
\label{tab:dataset}
\end{table*}
\section{System Descriptions}
\label{app:systems}
        \subsection{Ghostbuster}
        We use the implementation provided at \url{https://github.com/vivek3141/ghostbuster}. This was modified to fix an issue where outdated OpenAI models were referenced; we use \texttt{davinci-002} and \texttt{babbage-002} models.

        \subsection{Glimpse}
        We use the implementation provided at \url{https://github.com/baoguangsheng/glimpse}.
        
        \subsection{Binoculars \cite{hans-2024}} 
        We use the implementation provided at \url{https://github.com/ahans30/Binoculars}.

        \subsection{Fast-DetectGPT \cite{bao-2023}} 
        We use two settings that use different models for scoring: \texttt{gpt-neo-2.7b} for speed and \texttt{falcon-7b} for maximal accuracy.

        We use the implementation provided at \url{https://github.com/baoguangsheng/fast-detect-gpt}.

        \subsection{Zippy \cite{thinkst-2023}}
        We use the implementation provided at \url{https://github.com/thinkst/zippy}. We experimented with the LZMA and ensemble versions, and found no significant differences in performance.

        \subsection{BiScope \cite{guo-2024-1}} 
        We use the implementation provided at \url{https://github.com/MarkGHX/BiScope}: they do not provide an explicit "best" model for each domain, so we train each of our four variants using all the provided data from the respective domains.
        
        \subsection{DeTeCtivE \cite{guo-2024}} 
        We use the implementation provided at \url{https://github.com/heyongxin233/DeTeCtive}. 

        \subsection{RADAR \cite{hu-2023}} 
        We use the implementation provided at \url{https://github.com/IBM/RADAR}.

        \subsection{Desklib \cite{desklib-2025}}
        We use the implementation provided at \url{https://github.com/desklib/ai-text-detector}.

        \subsection{E5-lora \cite{dugan-2024}}
        We use the implementation provided at \url{https://github.com/menglinzhou/e5-small-lora-ai-generated-detector}. The creators indicate the desired citation is for the RAID dataset.

\subsection{Architecture/Costs}
\label{app:architecture}
For model training, inference, and evaluation we use Amazon AWS EC2 instances. We use the g6e.xlarge instance type. This instance type has an NVIDIA L40S Tensor Core GPU with 48 GB of GPU memory, allowing us to experiment with models that have larger GPU memory requirements (notably Binoculars and the FDG systems require significant GPU memory).

Running all models over our dataset requires approximately 6 hours of machine time, costing approximately \$12 USD. We additionally spent approximately \$200 USD for OpenAI model usage, required for the Ghostbuster and Glimpse models.
\section{Correlations}
\label{app:correlations}
\begin{figure}[t!]
\includegraphics[width=.49\textwidth]{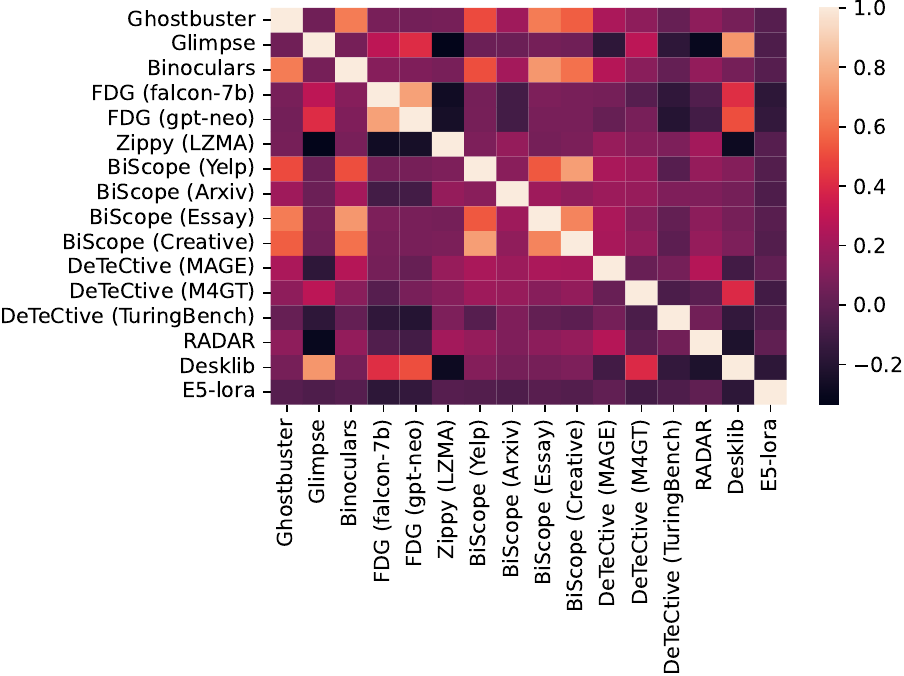}
\caption{Pearson correlation for the predictions for each model.}
\label{fig:corr} 
\end{figure}

Figure \ref{fig:corr} shows a heatmap of correlations between model predictions.

\section{Prompts}
\label{app:prompts}

We utilize the following prompt to interface with the language model (Claude 3.5). The prompt asks for completion of a given student essay. That essay is trimmed to the first 20 tokens, which are provided to the model with the instruction to complete the text. The model is instructed to limit the output to the length of the original essay, while mimicking the style of a student:

\begin{verbatim}
Here is the start of a student's essay. 
Complete the essay. It should be at 
most {len(text.split())} words long. 
Do not go over this requirement. 
Emulate the style of a student between 
6th and 12th grade. You may include 
some common misspellings and 
punctuation errors, so that the text 
looks like a students.
Start: 
{' '.join(text.split()[:20])}

Return only the resulting text as a 
json object:
{{\"text\":\"<generation>\"}} 
Ensure the result is under 
{len(text.split())} tokens."}
\end{verbatim}
\section{Annotators}
\label{app:annotators}

We recruited three annotators through personal requests for our experiments. These annotators are professionals with previous work in the field of deepfake detection, and consented to their results being used individually. All three annotators are post-graduate educated, fluent English speakers. Each annotator was given a batch of samples with the instruction to classify each sample as either human-written or machine-generated. Annotators were compensated as part of salaried work at a rate above minimum wage.

After completion of their individual tasks, we decided to involve the annotators as authors on the paper. Note that the annotators had no information about the project before their annotations were completed (which models, prompts, and datasets were used, what the goal of the project was, etc.): they were kept entirely independent until after their annotations were completed. This prevented any type of bias in having authors perform annotation. After annotations were completed, the annotators were able to make significant contributions to the analysis and writing of the paper, and given the amount of work and the level of contribution they made, we believed the best way to credit their work and contribution was to include them as authors.

\end{document}